\documentclass{article}

\PassOptionsToPackage{numbers, compress}{natbib}

\usepackage[preprint]{neurips_2021}

\usepackage[utf8]{inputenc} 
\usepackage[T1]{fontenc}    
\usepackage{hyperref}       
\usepackage{url}            
\usepackage{booktabs}       
\usepackage{amsfonts}       
\usepackage{nicefrac}       
\usepackage{microtype}      
\usepackage{mathtools} 
\usepackage{xcolor}         

\usepackage[linesnumbered,ruled,vlined,boxed]{algorithm2e}
\usepackage[inline]{enumitem}
\usepackage{graphicx}
\usepackage{rotating}
\usepackage{subcaption}
\usepackage{xspace}
\usepackage{bm}        
\usepackage{amssymb}   
\usepackage{adjustbox}
\usepackage{array}
\usepackage{subcaption}

\usepackage{multirow, wrapfig}
\usepackage{siunitx} 

\usepackage{amsthm}

\usepackage{xargs}                      
\usepackage[disable]{todonotes}
\newcommandx{\unsure}[2][1=]{\todo[linecolor=red,backgroundcolor=red!25,bordercolor=red,#1]{#2}}
\newcommandx{\change}[2][1=]{\todo[linecolor=blue,backgroundcolor=blue!25,bordercolor=blue,#1]{#2}}
\newcommandx{\info}[2][1=]{\todo[linecolor=yellow,backgroundcolor=yellow!25,bordercolor=yellow,#1]{#2}}
\newcommandx{\improvement}[2][1=]{\todo[linecolor=green,backgroundcolor=green!25,bordercolor=green,#1]{#2}}
\newcommandx{\thiswillnotshow}[2][1=]{\todo[disable,#1]{#2}}

\newcommand{\reffig}[1]{Fig. \ref{#1}}
\newcommand{\reftbl}[1]{Table \ref{#1}}
\newenvironment{horlist}{\begin{enumerate*}[label={\alph*)}]}{\end{enumerate*}}

\definecolor{matplotgreen}{RGB}{0, 128, 0}
\definecolor{matplotred}{RGB}{255, 0, 0}

\title{DAAIN: Detection of Anomalous and Adversarial Input using Normalizing Flows}

\author{%
  Samuel von Baußnern\thanks{equal contributions} \\
  \texttt{savoba@protonmail.ch} \\
  EPFL, Lausanne, Switzerland\\
  \And
  Johannes Otterbach${}^*$ \\
  \texttt{johannes.otterbach@merantix.com} \\
  Merantix Labs GmbH, Berlin, Germany\\
  \And
  Adrian Loy\\
  \texttt{adrian.loy@merantix.com} \\
  Merantix Labs GmbH, Berlin, Germany\\
  \And
  Mathieu Salzmann\\
  \texttt{mathieu.salzmann@epfl.ch} \\
  EPFL, Lausanne, Switzerland\\
  \And
  Thomas Wollmann\\
  \texttt{thomas.wollmann@merantix.com} \\
  Merantix Labs GmbH, Berlin, Germany\\
}

\begin{document}

\maketitle

\begin{abstract}
Despite much recent work, detecting out-of-distribution (OOD) inputs and adversarial attacks (AA) for computer vision models remains a challenge. In this work, we introduce a novel technique, DAAIN, to detect OOD inputs and AA for image segmentation in a unified setting. Our approach monitors the inner workings of a neural network and learns a density estimator of the activation distribution. We equip the density estimator with a classification head to discriminate between regular and anomalous inputs. To deal with the high-dimensional activation-space of typical segmentation networks, we subsample them to obtain a homogeneous spatial and layer-wise coverage. The subsampling pattern is chosen once per monitored model and kept fixed for all inputs. Since the attacker has access to neither the detection model nor the sampling key, it becomes harder for them to attack the segmentation network, as the attack cannot be backpropagated through the detector.
We demonstrate the effectiveness of our approach using an ESPNet trained on the Cityscapes dataset as segmentation model, an affine Normalizing Flow as density estimator and use blue noise to ensure homogeneous sampling. Our model can be trained on a single GPU making it compute efficient and deployable without requiring specialized accelerators.
\end{abstract}

\section{Introduction}

Deep learning has led to great breakthroughs in various tasks, such as image recognition \cite{heDeepResidualLearning2016}, speech processing \cite{hintonDeepNeuralSpeech2021} and robotics training \cite{levineEndToEnd2016}. As deep neural networks are are being deployed in real world settings it becomes increasingly important to understand model uncertainties and failure modes. This is especially true in high stake scenarios such as autonomous driving or breast cancer detection \cite{eykholt2018robust, lotterRobustBreastCancer2021}. In this work we focus on failure modes originating from out-of-distribution (OOD) inputs and adversarial attacks (AA) to the model. We expand the existing toolset of AA and OOD detection to the case of image segmentation models \cite{ronneberger2015unet, Chaurasia_2017, zhao2017pyramid, lin2017feature, chen2017rethinking}, which has only received little attention so far \cite{nakka2020indirect, lis2019detecting, xiao2018characterizing}.

Specifically, we use Normalizing Flows \cite{Kobyzev_2020, dinhNICENonlinearIndependent2015} in conjunction with blue noise sampling \cite{bridsonFastPoissonDisk2007} to estimate the density of a random but fixed set of the model's internal activations. Thanks to our density estimation approach we do not need access to an OOD or adversarial distribution at training time since activations of irregular inputs will be mapped to different regions in latent space than the regular ones, corresponding to lower log-likelihoods \cite{bishop1994novelty}. Despite recent concerns of learning reliable density estimators \cite{choiWAICWhyGenerative2019, lanPerfectDensityModels2020}, we show that this approach performs surprisingly well in practice. After mapping the activations to the latent space, we apply an additional classifier to discriminate between regular and anomalous inputs. This allows us to monitor the model's behavior, making it resilient even to white box attacks if the attacker is prevented from accessing the detection model activations and/or the blue noise sampling key.

We evaluate our method on the Cityscapes dataset \cite{cordtsCityscapesDatasetSemantic2016} and significantly outperform baselines adopted from image classification studies that similar to our approach also apply to both OOD and AA detection. Importantly, our model can be trained and tested on a single GPU, making it resource efficient and deployable in an accelerator-free environment. Our main contributions can be summarized as:
\begin{itemize}
    \item We investigate adversarial and OOD attacks against image segmentation models.
    \item We use Normalizing Flows to map a random but fixed set of activations to a latent space where we can estimate log-likelihoods and train a classifier for OOD and AA detection.
    \item We introduce blue noise sampling to homogeneously sample the full activation space, facilitating scalable anomaly detection.
    \item Our detection model can be trained with limited compute resources and can be deployed without special hardware accelerators.
\end{itemize}

While we investigate and discuss our detection scheme for image segmentation, we believe our approach to readily generalize to other image-related tasks and data modalities. Moreover, the density estimator is not limited to Normalizing Flows and could be replaced with other representation learning architectures. The code for this work can be found at \texttt{\href{https://github.com/merantix/mxlabs-daain}{https://github.com/merantix/mxlabs-daain}}.

\begin{figure}[t]
  \centering
  \includegraphics[width=.8\linewidth]{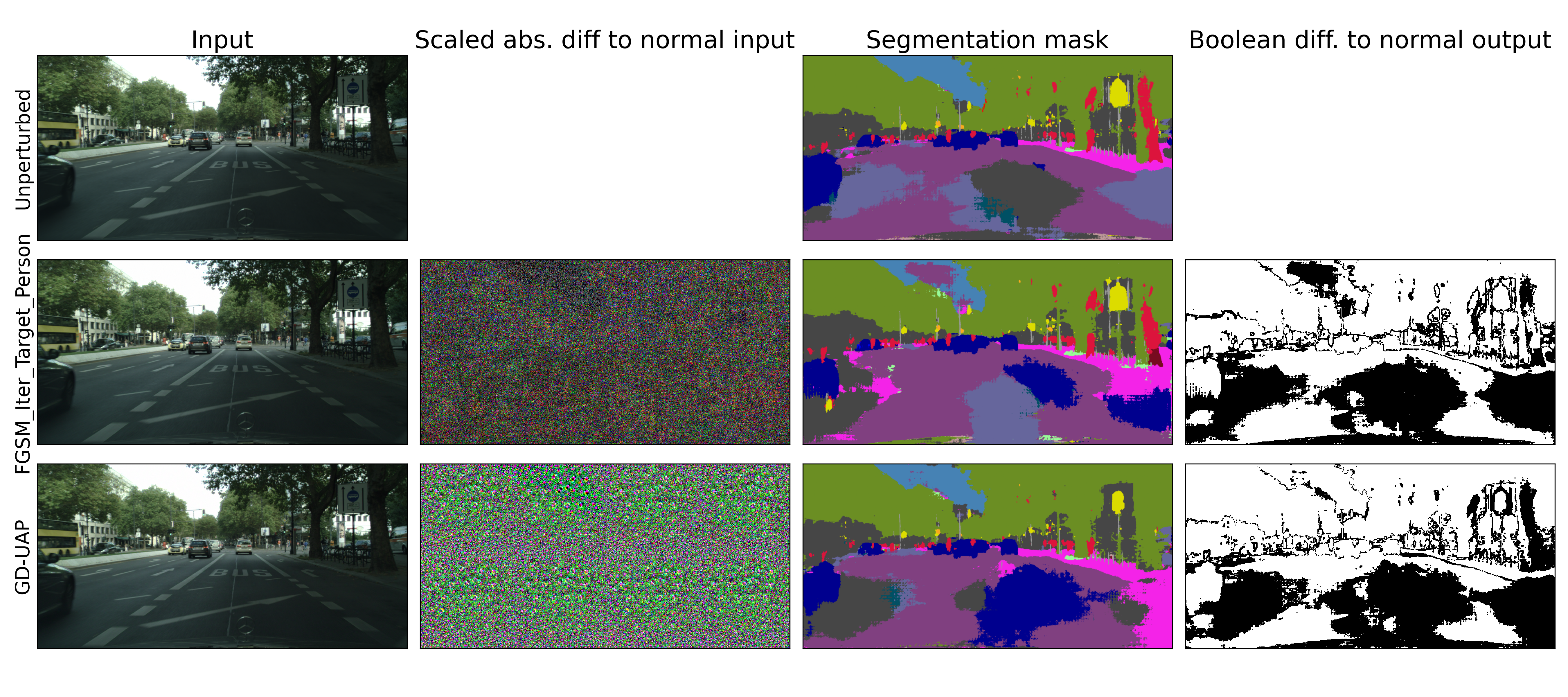}
  \caption{Example perturbations used in this work. The first row shows the normal, unperturbed input, and the next two rows show two example attacks or perturbations. The columns show the input, the difference to the normal input upscaled  within $[0, 255]$ for visualisation purposes, the output of the segmentation model, and a binary mask depicting if the attack was successful for any pixel.}
  \label{fig:small_perturbations_example}
\end{figure}

\begin{figure}[t]
  \centering
  \includegraphics[width=.7\linewidth]{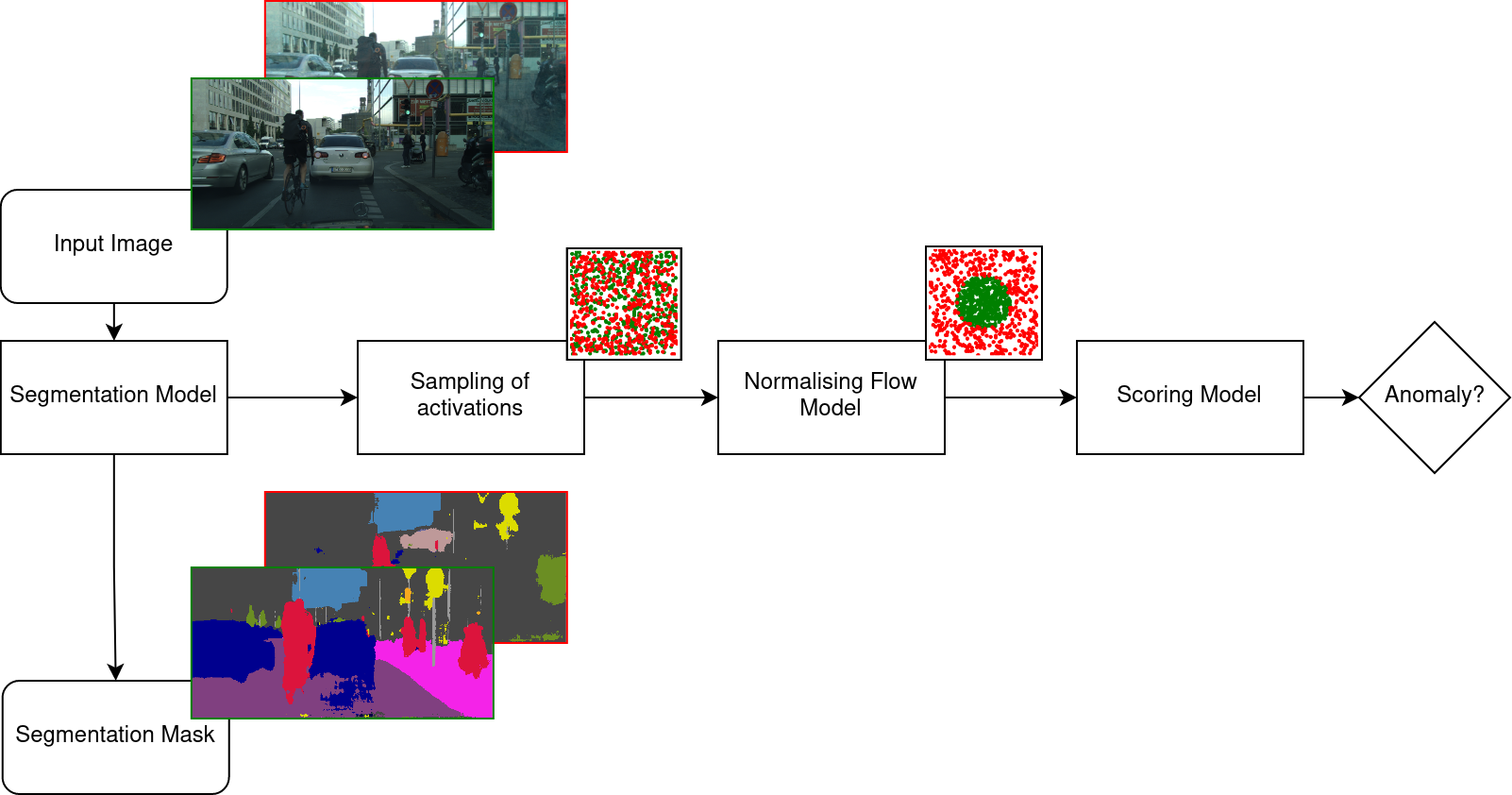}
  \caption{Sketch of a full deployment including the segmentation and detection models. The regular and anomalous inputs as well as the respective activations are marked \textcolor{matplotgreen}{green} and \textcolor{matplotred}{red}. The activations and transformed activations are plotted in the same image for presentation purposes. The transformed activations in the latent space of the Normalizing Flow indicate the ideal case where the anomalous input is mapped far away from the point of origin.}
  \label{fig:pipeline}
\end{figure}

\section{Related Work}\label{sec:related_work}

\textbf{Classification.} Image recognition tasks are widely used in real-world deployments, making them a high-priority setting for detecting OOD and AA inputs. E.g., Hendrycks at al.~\cite{hendrycksBaselineDetectingMisclassified2018} develop simple baselines for OOD detection using the maximum softmax prediction (MSP) score of a classifier. They demonstrate the effectiveness of this baseline across a range of modalities. Metzen et al.~\cite{metzenDetectingAdversarialPerturbations2017} follow a supervised approach to train small convolutional networks on activations extracted from a pretrained ResNet to distinguish regular from adversarial activation patterns. Roth et al.~\cite{rothOddsAreOdd2019} study the effect of noise on top of regular and adversarial inputs in the context of image classification. They show that changes in log-odds have specific characteristics depending on their regularity under $L^\infty$-attacks and exploit this fact to build an AA detector. Grosse et al.~\cite{grosseStatisticalDetectionAdversarial2017} show that standard statistical models are able to exploit distributional differences between regular and anomalous inputs with as few as 50 samples to detect AA in image classification. Moreover they show that including an additional ``adversarial" output class during training enables the model to learn the distributional features of adversarial input and defend against those. Qin et al.~\cite{qinDetectingDiagnosingAdversarial2020} investigate the success rate of AAs in the context of Capsule Networks and show that these networks are learning features more aligned with the visual perceptions of humans by demonstrating that an adversarial attack produces images more closely related to the visually accepted class. Lee et al.~\cite{leeSimpleUnifiedFramework2018} use a class conditional Gaussian discriminant analysis based on high- and low-level features from a pretrained network to detect OOD and AA input in a single framework. The Mahalanobis distance is used as a measure of how confident a model is to assign a given class. In~\cite{rudolphSameSameDifferNet2020}, Rudolph et al. use Normalizing Flows to estimate the density of features extracted from the convolution operations of AlexNet~\cite{alexNet} for defect detection. They construct their anomaly score based on the average log-likelihood of the original and perturbed images. Feinman et al.~\cite{feinmanDetectingAdversarialSamples2017} employ Gaussian kernel density estimates trained on the last hidden layer of a classification network in conjunction with Bayesian uncertainty estimates based on dropout for discriminating between noisy and adversarial samples constructed from original images.

\textbf{Segmentation.} In this work, we focus on OOD and AA detection for segmentation models. In this context, Xia et al.~\cite{xiaSynthesizeThenCompare2020} and Lis et al.~\cite{lis2019detecting} use a generative approach to reconstruct the input based on the predicted segmentation mask and compare it to the original input. They show that large differences in the reconstructions correspond to large model uncertainty. The success of this method strongly depends on the decoder's capabilities, making it harder to generalize to new settings and potentially compute intensive. Rottmann et al.~\cite{rottmannPredictionErrorMeta2019} build a meta-classifier to predict the reliability of a predicted segment in an image. They train a logistic classifier based on hand-engineered aggregated metrics derived from intersection-over-union (IOU) measures between predicted mask and ground truth and show that those metrics contain significant information about the reliability of the segmentation predictions. Defard et al.~\cite{defardPaDiMPatchDistribution2020} extract pixel-wise feature vectors from pretrained CNNs to distinguish regular from anomalous input in a one-class setting to detect manufacturing anomalies. They construct embeddings for the feature vectors and use multivariate Gaussian estimates to determine the irregularity of a new feature vector. To apply this to anomaly detection to an input image, the task is done patch-wise, rather than on the whole input image, to address computational and dimensionality concerns. Their setting is tailored to manufacturing anomalies, which are distinct from the OOD and AA distributions we consider in this work. Xiao et al.~\cite{xiao2018characterizing} develop an AA detector for segmentation tasks based on a spatial consistency score. They show that the mean IOU between two randomly selected patches is higher in regular input and lower for AA. Exploiting these statistical differences, they construct a reliable detector against adversarial input, but do not study the case of OOD inputs. Nakka and Salzmann~\cite{nakka2020indirect} investigate localized adversarial attacks that are intended to interfere with an object of interest outside the attacked region. They show that the global context encoded in segmentation networks renders these attacks highly effective and propose a detection scheme based on a pixel-wise score constructed by fitting a class-conditional Gaussian kernel to the inner activations of the segmentation model. AA detection is done using the Mahalanobis distance of a new sample to the class centers. The effectiveness of this approach was not demonstrated on OOD inputs. Ren et al.~\cite{renLikelihoodRatiosOutofDistribution2019} show that background statistics have a significant impact when using the likelihood score of generative models for OOD detection. They propose a likelihood-ratio method to correct for this and study the method using autoregressive models on a newly created genomic sequence dataset, while their image datasets are restricted to small-scale MNIST and CIFAR-10 inputs. These models are not specific to classification or segmentation and would need to be trained and deployed in addition to the task-specific network and their robustness to OOD or AA input is unclear.

While previous works studied the impact of OOD or AA inputs in the context of image segmentation, they never address both in a unified framework, as has been done in the classification setting. Moreover, some approaches  target special purpose setups (manufacturing, hand-engineered features) or require powerful generative decoders to work, typically resulting in high compute needs. In contrast, we address these limitations in a resource-efficient, unified framework that applies to both OOD and AA inputs for full-size image segmentation models, and has the potential to generalize to other datasets and modalities.

\section{Model Design}\label{sec:model_design}

Our detection system major components are the segmentation model, the density estimator of the sampled activations and the classification head (see \reffig{fig:pipeline}). We use an ESPNet pretrained on Cityscapes \cite{mehtaESPNetEfficientSpatial2018} as segmentation model, which we keep frozen during all experiments. We choose affine Normalizing Flows with ActNorm \cite{kingmaGlowGenerativeFlow2018} as density estimators and train them on a subsampled distribution of the ESPNet's activations. In an ideal scenario, the network activations are already close to a Gaussian and expect the Normalizing Flow does not need to be overly expressive to disentangle the remaining signal in the activations into useful representations to train a simple classification head for the detection task. Moreover, the benefit of using a Normalizing Flow is its lossless compression that preserves the mutual information on the input as well as latent side by construction \cite{tschannen2020mutual}, while the latent space is encouraged to respect the Gaussian assumption necessary to reliably train a logistic regression classifier \cite{hastie_09_elements-of-statistical-learning}.

\subsection{Subsampling Procedure}

Na\"{i}vely, training a Normalizing Flow on the full distribution of activations in a neural network is not scalable due to the number of parameters in such a network. To address the high dimensionality of the activation space we employ dimensionality reduction techniques. As shown in \cite{qiuSemanticAdvGeneratingAdversarial2020, nakka2020indirect} the properties of convolutional networks allow attacks that are constrained to a small region in the input to propagate outside this region in a downstream classification and/or segmentation task. We hypothesize that this propagation effect is observable in the activation space as well and that, by sampling and observing activations that in pixel-space are far removed from the attacked patch, we can still detect such local attacks. This allows us to subsample the activation space to reduce dimensionality. However, any structured subsampling strategy would give rise to information the attacker can exploit to circumvent the attack and hence we use a random sampling technique. On the other hand, pure random sampling still requires us to sample a large number of activations to train a reliable monitor model. Therefore, we choose an intermediate strategy using blue noise sampling \cite{bridsonFastPoissonDisk2007} to have a homogeneous coverage of the activation space while still being random due to the initial seed of the sampling procedure. Specifically, we first record all inner activations and then upscale them to the input dimensionality using nearest neighbor upsampling. The resulting 3D volume is sampled according to a fixed blue noise sampling key, resulting in fixed sampling locations for the detection model.

\begin{figure}[t]
  \centering
  \includegraphics[width=.8\textwidth]{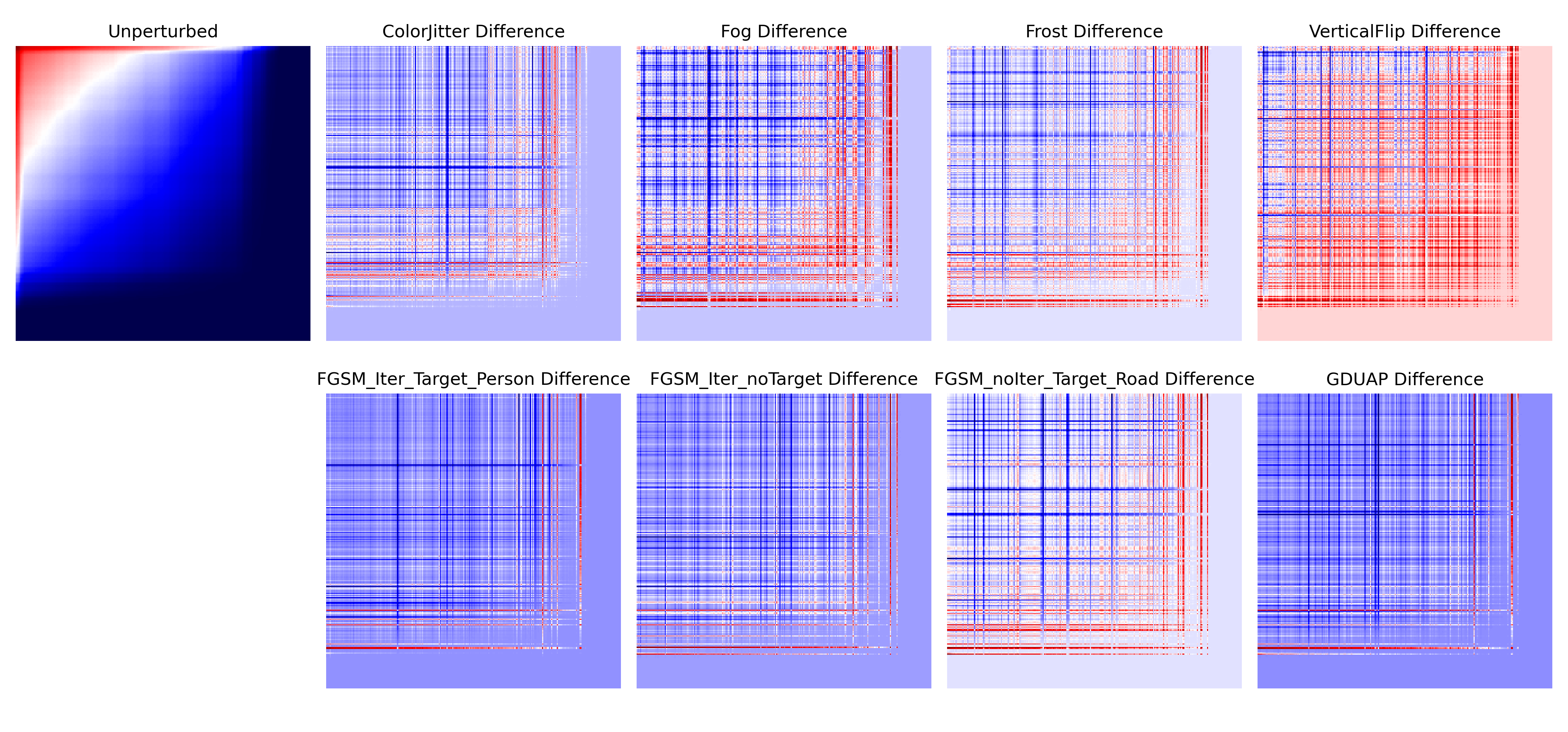} 
  \caption{Coactivations of the upscaled and flattened activations for a single image over multiple perturbations. The coactivations are the pairwise product of the activations and were log-transformed for presentational purposes. For visibility, we order the co-activations according to their strength in the unperturbed case and show only the difference in the co-activations between a specific perturbation and the unperturbed baseline. We can clearly see how different perturbations result in very different but distinct coactivation patterns.}
  \label{fig:espnet_coactivations_cityscapes}
\end{figure}

To show that such a random sampling is sufficient to build a successful detection model, we look at the coactivation correlations for various unperturbed, OOD and AA inputs. The results in \reffig{fig:espnet_coactivations_cityscapes} show the difference induced by the perturbations compared to the background coactivations of the unperturbed input. We can clearly see that the perturbations switch off certain coactivations while activating others and hypothesize that the irregular input tries to exploit dead neurons, which are not activated with in-distribution inputs. This suggests that learning a faithful density estimator of the activation space of the regular input should be sufficient to train a reliable detection model that can distinguish between different input perturbations.

\subsection{Density Estimation of the Activation Space}
As indicated in the previous section, we should be able to detect anomalous changes in the input by monitoring changes in a random but fixed subset of the activation space. To maximize the utility of the sampled latent space, we aim to learn a set of meaningful representations while preserving the mutual information between the activation and representation space. To this end we choose Normalizing Flows as a framework for density estimation, as they have been shown to learn disentangled representations \cite{sorrensonDisentanglementNonlinearICA2020} and preserve the mutual information by construction \cite{tschannen2020mutual}.

Normalizing Flows are a class of density estimators that use a repeated application of the change-of-variable theorem to transform a complex input distribution $p(\bm{x})$, $\bm{x}\in\mathbb{R}^D$ into a Gaussian latent space according to
\begin{equation}
    p(\bm{x}) = p(\bm{z}) \left\vert\det\frac{\partial \bm{z}}{\partial \bm{x}} \right\vert\;,
\end{equation}
where $\frac{\partial \bm{z}}{\partial \bm{x}} = \left(\frac{\partial z_i}{\partial x_j}\right)_{i,j \in\{1, \cdots, D\}}$ denotes the Jacobian of the transformation \cite{Kobyzev_2020}. The computation of the Jacobian determinant is costly and there has been a flurry of works proposing techniques to design expressive transformations while reducing the computation cost \cite{dinhDensityEstimationUsing2017, kingmaGlowGenerativeFlow2018, pmlr-v97-behrmann19a, papamakarios2018masked, ho2019flow, grathwohl2018ffjord}. In this work we restrict ourselves to the use of affine transformations \cite{dinhDensityEstimationUsing2017}. We split the input $\bm{x} = (\bm{x_1}, \bm{x_2})$ into two components $\bm{x_1} = (\bm{x}_i)_{i \in\{1, \cdots, d\}}$ and $\bm{x_2} = (\bm{x}_i)_{i\in \{d+1, \cdots, D\}}$ and define an element-wise affine operation according to
\begin{equation}\label{eq:affine_coupling_block}
\begin{aligned}
   \bm{z}_1 & = \bm{x}_1 \\
   \bm{z}_2 & = \bm{x}_2 \odot s(\bm{z}_1) + t(\bm{z}_1)\;,
\end{aligned}
\end{equation}
where $\odot$ denotes element-wise multiplication. Due to the design of the coupling block, the Jacobian is lower-triangular and the determinant is readily computed. The above transformation preserves the ordering in the input-output relation, and we need to introduce coordinate mixing to condition each variable on all others using $1\times 1$\,-convolutions, attention mechanisms or dense permutation layers. As usual in density estimation, it is advantageous to work in log-space and minimize the negative log-loss
\begin{equation}
    \mathcal{L}(\bm{x}) = -\log p(\bm{x}) = -\log p(\bm{z}) - \log\left\vert\det\frac{\partial \bm{z}}{\partial \bm{x}} \right\vert\;.
\end{equation}

\begin{figure}[t]
  \centering
  \includegraphics[width=\linewidth]{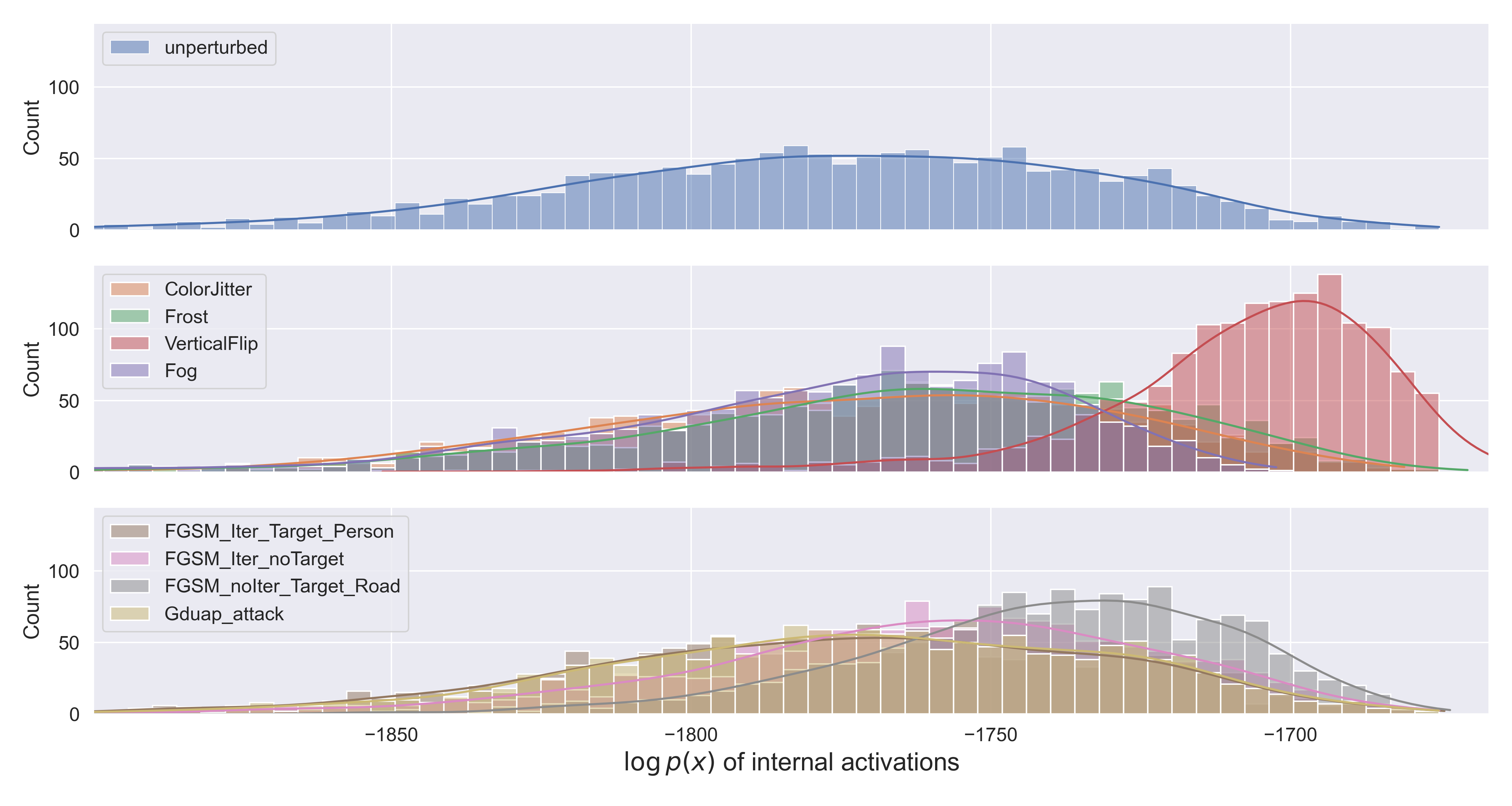}
  \caption{Histogram of $\log p(\bm{x})$ for the unperturbed test activations (top) as well as OOD modifications (middle) and adversarial attacks (bottom) using a GIN Normalizing Flow \cite{sorrensonDisentanglementNonlinearICA2020} with $D$=1784 dimensions as a density estimator. While the flow fails to assign higher likelihoods to the OOD and AA inputs \cite{nalisnickDeepGenerativeModels2019} the distributions show significant differences such that an additional classification head can be used to differentiate between regular and anomalous inputs.}
  \label{fig:p_z_test}
\end{figure}

In \reffig{fig:p_z_test} we show the log-likelihood density estimates of the unperturbed samples of the test dataset as well as the anomalous input after training a Flow on a training set constructed by subsampling activations over the unperturbed training split of the Cityscapes dataset (see the experiments section for more details). As already observed in \cite{nalisnickDeepGenerativeModels2019} the trained flow fails to assign lower likelihoods to OOD as well as AA inputs, rendering detection by simple density estimation impossible. However, one can also see that  the various OOD and AA inputs have different densities indicating that a classification head on top of the latent representations should be able to differentiate between the regular and anomalous inputs.

Before moving on to the experimental results, it should be noted that our approach does not critically rely on the use of Normalizing Flows but can be generalized to use any other density estimation or representation learning algorithm to circumvent the potentially-flawed density estimates of Normalizing Flows~\cite{nalisnick2019deep, kirichenko2020normalizing}. Specifically, we do not rely on the density estimate of the flow to determine anomalous inputs, but rather use it as a feature extractor for a classification head. It is well known that Normalizing Flows are diffeomorphisms and as such preserve topological invariants \cite{cornish2021relaxing, huang2020augmented}, limiting their ability to learn disentangled representations. However, our subsampling procedure breaks such topological constraints in the input space, enabling the flow network to learn reasonably disentangled representations \cite{wu2020stochastic}.

\subsection{Classification Head}
To keep things simple, we focus on a non-trainable classification head, although it is straight-forward to include a logistic regression head or any other kind of trainable linear classifier. This classification head acts directly on the latent variables and can be expressed through various distance metrics $\tau$. Formally the classification is given by
\begin{equation}
\mathcal{C}_\theta(\bm{x}) =
\begin{cases}
  1 & \text{if } \tau(z(\bm{x})) \geq \theta \\
  0 & \text{otherwise}
\end{cases}
\label{eq:distance_classifier}
\end{equation}
with $\mathcal{C}_\theta(\bm{x}) = 1$ indicating an anomaly and $z(\bm{x})$ being the latent code corresponding to the input picture $x$ as defined by the Normalizing Flow's change-of-variable transformation. The threshold $\theta$, defining the classification boundary, needs to be determined manually, but subsequent work could focus on automating this process.

Possible choices for the distance kernel $\tau$ are 
\begin{horlist}
\item the Euclidean distance $\tau_{\text{Euclidean}}(\bm{z}) = \frac{1}{n} \sum_{i = 1}^n z_i^2$, which (up to constants) corresponds to the negative log-likelihood of the latent variables under the latent prior of the flow;
\item the harmonic mean distance $\tau_{\text{Harmonic}}(\bm{z}) = n\left(\sum_{i = 1}^n z_i^{-2}\right)^{-1}$;
\item the Malahanobis distance 
$
\tau_{\text{Mahalanobis}}(\bm{z}) = \sqrt{
  (\bm{z} - \bm{\mu}_\text{train})^T
  \bm{S}_\text{train}^{-1}
  (\bm{z} - \bm{\mu}_\text{train})
}
$, based on a Gaussian kernel estimate with centroid $\bm{\mu}_\text{train}$ and covariance $\bm{S}_\text{train}$ as defined, e.g., in \cite{leeSimpleUnifiedFramework2018}, estimated from the training data distribution; or
\item a Histogram-Based Outlier Score (HBOS) $\text{HBOS}_k(\bm{z}) = \sum_{i = 1}^n \log \left(\frac{1}{\text{hist}_i(z_i)}\right)$, where each $\text{hist}_i$ is a fitted histogram for the dimension $i$ of the given input with $k$ bins as a hyper-parameter.
\end{horlist}

\section{Experiments}\label{sec:exp}
We use a pretrained segmentation model, based on an ESPNet architecture and trained on the Cityscapes dataset \cite{mehtaESPNetEfficientSpatial2018, cordtsCityscapesDatasetSemantic2016}. The dataset consists of $3475 = 2975\ (\text{training}) + 500\ (\text{validation})$ fully segmented RGB images with a resolution of $2048 \times 1024$ pixels taken in 44 different cities throughout Europe. The images are annotated using 19 classes and one background or void class. We downsample all images and labels to a resolution of $512 \times 256$ pixels using bilinear and nearest-neighbour interpolation for images and labels, respectively, to increase experimental iteration speed. The full training data was used to train the detection model, and the full test split was further used for the attack and anomaly generation.

\begin{table}[t]
{\small
\begin{subtable}[align=top]{0.5\textwidth}
    \begin{tabular}{lrr}
    \toprule
    Perturbation &  AAcc &  AIoU \\
    \midrule
    ColorJitter             & 0.135 & 0.624 \\
    Fog                     & 0.848 & 0.974 \\
    Frost                   & 0.301 & 0.760 \\
    VerticalFlip            & 0.882 & 0.964 \\
    FGSM\_Iter\_Target\_Person & 0.313 & 0.804 \\
    FGSM\_Iter\_noTarget      & 0.728 & 0.930 \\
    FGSM\_noIter\_Target\_Road & 0.752 & 0.935 \\
    GD-UAP            & 0.359 & 0.816 \\
    \bottomrule
    \end{tabular}
\end{subtable}
\begin{subtable}[align=top]{0.5\textwidth}
    \begin{tabular}{lrrr}
    \toprule
    Scoring Function &  AUROC &  AUPR & FPR95 \\
    \midrule
    Euclidean &  0.673 & 0.662 & 0.680\\
    HarmonicMean &  0.548 & 0.557 & 0.934 \\
    $\text{HBOS}_{30}$   &  0.723 & 0.705 & 0.616 \\
    $\text{HBOS}_{40}$   &  0.723 & 0.705 & 0.615\\
    $\text{HBOS}_{20}$   &  0.722 & 0.705 & 0.617\\
    Mahalanobis &  0.567 & 0.593 & 0.869 \\
    \bottomrule
    \end{tabular}
\end{subtable}
}
\caption{(\textit{left})Attack Success Rate measured as the average attack accuracy per pixel (\textsc{AAcc}) and average intersubsection-over-union (\textsc{AIoU}) \cite{nakka2020indirect, metzenDetectingAdversarialPerturbations2017, xieAdversarialExamplesSemantic2017}. (\textit{right}) Comparison of results for scoring function averaged over all perturbations. We chose the highest result over all model-configuration ablations. The choice of the classification function is highly sensitive, leading to strong variations in performance.}
\label{tab:attack_succes_rate_and_clf_head_perf}
\end{table}

\subsection{Baselines and evaluation metrics}
We compare the method to two baselines used in \cite{hendrycksBaselineDetectingMisclassified2018} for anomaly detection for image classification. The first one is based on the maximum softmax prediciton (MSP). Using the classification model $f_s: \mathbb{R}^{h \times w \times c} \longrightarrow \mathbb{R}^{h \times w \times k}$, we assign the MSP score as $\text{MSP}(f_s, \bm{x}) = \max(f_s(\bm{x}))$. At test time we assign binary labels to the input according to it being regular or anomalous. We record the maximum MSP score over all pixels in the input and treat it as a score predicting the binary label. The performance of this predictive score model can be assessed through various binary classifier metrics, such as AUROC, AUPR, FPR.

The second baseline is an extension of this idea using Monte-Carlo dropout (MCD), introduced in \cite{galDropoutBayesianApproximation2016}, to simulate a Bayesian approximation to a Gaussian process. We introduce dropout layers into the pretrained ESPNet in front of all weight-layers and fine-tune the ESPNet for another 30 epochs with a dropout rate of 0.1 using the original data-augmentation and training setup of \cite{cordtsCityscapesDatasetSemantic2016}. To turn this fine-tuned dropout segmentation model into an anomaly detector we enable dropout at test time and feed the input 10 times through the segmentation model. We record the pixel-wise MSP for each feed-forward pass and calculate the variance over the score irrespective of the predicted class. Intuitively, this score should capture the uncertainty in the model due to an unexpected and anomalous input. To predict anomalous input, we proceed like in the MSP case, by using the maximum MSL variance over the whole image as a predictive score for irregularity. To evaluate the model and the baselines we treat anomalous and adversarial examples as the positive class, and use three metrics:
\begin{horlist}
\item the area under the receiver operating characteristic curve (AUROC),
\item the area under the precision-recall curve (AUPR) and
\item the false positive rate at $N$\% true positive rate (FPRN)
\end{horlist}. When evaluating the metrics, we compare the detection model on a single perturbation at a time, i.e., we use the full unperturbed test set and choose a single perturbation for the whole test set to determine a threshold $\theta$ of the distance based classifier \eqref{eq:distance_classifier}. This means that the threshold could vary between different perturbations and follow-up work should focus on removing this limitation and extend the method to work on all perturbations at the same time.

\subsection{Training setup of the flow-based detection model}\label{sec:training_setup}
We implemented the Flow model using the FrEIA framework \cite{freia-framework}. To train the model we sample the activations based on a minimum distance of 128 and 64 for the blue noise samples, resulting in a flow of fixed dimensionality $D=$262 and 1784, respectively. Note that given the stochasticity of the sampling mask creation process this number varies from run to run. 

We train the model on a single GTX-1070 GPU using PyTorch 1.7.1 with CUDA version 11.2 \cite{NEURIPS2019_9015}. We use the train-test split of Cityscapes used in \cite{mehtaESPNetEfficientSpatial2018} and employ the Adam optimizer \cite{kingmaAdamMethodStochastic2017} with a learning rate of 0.0002, beta value of 0.8 and weight decay of $10^{-5}$. To simulate the real-world setting where new attacks and anomalies are not know prior to deployment, no additional training data was used. Furthermore no augmentations were used. To mitigate training instabilities we use soft gradient-clamping of 0.5. The results are averaged over five runs with differently initialized weights. The models were trained with early stopping and no upper limit with respect to the number of epochs. The weights are initialized with the default initialization method; Xavier normal initialization did not show an huge impact. The batchsize is chosen to be 64 (32) for the setup using a minimum distance of 128 (64) for the blue noise. We use the full train split to train the detection model and the full test set to generate anomalous inputs and construct the anomaly detection head.

To generate the anomalies we follow \cite{hendrycksBenchmarkAnomalySegmentation2019} and apply vertical flipping transformations as well as fog and frost mask corruptions to the original data with the default parameter settings. For generating adversarial attacks we focus on gradient based white-box attacks to test the framework against the strongest possible adversaries. Specifically we use \begin{horlist}
\item non-targeted, non-data-depended, transferable attacks using \texttt{GD-UAP} \cite{mopuriGeneralizableDatafreeObjective2018a} with a \texttt{DL-VGG16} base model,
\item iterated FGSM \cite{kurakinAdversarialExamplesPhysical2017, goodfellowExplainingHarnessingAdversarial2015} with the target class being a person (\texttt{FGSM\_Iter\_Target\_Person}) and hyperparameters $\alpha=$0.005, $\epsilon=$0.01 and n\_iters=5
\item iterated FGSM with no target, i.e. the target is the background class (\texttt{FGSM\_Iter\_noTarget}). Hyperparameters are the same as before
\item simple FGSM \cite{goodfellowExplainingHarnessingAdversarial2015} (\texttt{FGSM\_noIter\_Target\_Road}) with the target class being the road and hyperparameters being $\alpha=$0.005, $\epsilon=$0.02.
\end{horlist} These settings generate attacks of various degree while still having a measurable effect on the segmentation model's performance, evaluated by the average attack accuracy per pixel (\textsc{AAcc}) and average intersubsection-over-union (\textsc{AIoU}) between the original output and perturbed output (cf. \reftbl{tab:attack_succes_rate_and_clf_head_perf} left). For visual examples of the attacks and their effects see \reffig{fig:small_perturbations_example}.

\subsection{Results}\label{sec:results}

\begin{table}[t]
{\small
\begin{tabular}{lllllllrrrrrr}
\toprule
ST & CB & SL & MD & SF & AUROC &  AUPR &  FPR95  & NP & FLOP & IT [ms] \\
\midrule
linear      & 16                & EW        & 64                 & $\text{HBOS}_{30}$ & 0.753 & 0.742 &  0.580 & 5.1e5 & 5.1e7 & 102.6\\
linear      & 16                & EW        & 64                & $\text{HBOS}_{40}$ & 0.753 & 0.742 &  0.579 & 5.1e5 & 5.1e7 & 102.6\\
linear      & 16                & EW        & 64                & $\text{HBOS}_{20}$ & 0.752 & 0.741 &  0.579 & 5.1e5 & 5.1e7 & 102.6\\
linear      & 8                 & EW        & 64                 & $\text{HBOS}_{20}$ & 0.752 & 0.746 &  0.578 & 2.6e5 & 2.6e7 & 55.6\\
linear      & 8                 & EW        & 64                 & $\text{HBOS}_{30}$ & 0.752 & 0.746 &  0.578 & 2.6e5 & 2.6e7 & 55.6\\
\bottomrule
\end{tabular}
}
\centering
  \caption{Overall results of the five best performing models based on the average AUROC over all perturbations and the corresponding hyperparameter settings and compute requirements. For the different values we refer to the main text. Inference time (IT) was measured at batchsize 1 with no gradient accumulation on an Intel Core i7 8th Gen averaged over 10 runs (ST: subnet-type, CB: number of coupling blocks, SL: sampling locations, MD: minimum distance for blue noise, SF: scoring function, NP: number of trainable parameters, IT: inference time, EW: everywhere).}
  \label{tab:overall_results}
\end{table}
We perform hyperparameter ablations over different configurations of the activation sampling procedure (minimum distance for blue noise: 64, 128; sampling locations: everywhere (EW), output of convolutional layer, before batchnorm layers), flow-types (mixing-layers: $1\times 1$\,-convolution, linear layers; GIN \cite{sorrensonDisentanglementNonlinearICA2020} coupling: yes, no; number of blocks: 4, 8, 16) and distance metrics (Euclidean, harmonic mean, Mahalanobis, HBOS${}_{20}$, HBOS${}_{30}$, HBOS${}_{50}$) and compare the results  to the baseline models. The results for the best performing model for each metric and each task are reported in \reftbl{tab:performance}, clearly showing that DAAIN outperforms the baselines on the majority of tasks. Note that the best performing configurations do not necessarily correspond to the configurations used to produce the log-likelihood visualizations in \reffig{fig:p_z_test}. To highlight the low compute requirements we list the hyperparameter configurations and the corresponding compute requirements for the top five best performing models based on the average AUROC over all perturbations in \reftbl{tab:overall_results}.

As shown by the attack success rates of \reftbl{tab:attack_succes_rate_and_clf_head_perf} (left), not every attack or perturbation is equally effective. Detection of Fog, VerticalFlip, FGSM\_Iter\_noTarget and FGSM\_noIter\_Targeti\_Road is of higher importance, given the higher impact of these perturbations on the segmentation model. The proposed method generally performs better than the baselines, which is to be expected given the fact that those where not designed to the task. Unfortunately, the overall high false-positive rate might prohibit a direct real-world application.

Finally, the choice of scoring function is important as seen in \reftbl{tab:attack_succes_rate_and_clf_head_perf}(right). This should come as no surprise given the discussion about the shortcomings of Normalizing Flows and the construction of the classification head. It indicates that we can expect further improvements by replacing the parameter-free head with a simple logistic regression head or other linear classifiers.

\begin{table}[t]
  \centering
\adjustbox{max width=\textwidth}{
\begin{tabular}{lccccccccc} 
\toprule
                           & \multicolumn{3}{c}{AUROC} & \multicolumn{3}{c}{AUPR} & \multicolumn{3}{c}{FPR95}  \\
Perturbation               & DAAIN  & MCD   & MSP       & DAAIN  & MCD   & MSP      & DAAIN  & MCD   & MSP        \\ 
\midrule
ColorJitter                & \textbf{0.667} & 0.418 & 0.500     & \textbf{0.665} & 0.459 & 0.500    & \textbf{0.860} & 0.988 & 1.000      \\
Fog                        & \textbf{1.000} & 0.777 & 0.424     & \textbf{1.000} & 0.826 & 0.465    & \textbf{0.001} & 0.882 & 0.984      \\
Frost                      & 0.850 & 0.598 & \textbf{0.000}     & \textbf{0.857} & 0.603 & 0.307    & \textbf{0.591} & 0.902 & 1.000      \\
VerticalFlip               & \textbf{1.000} & 0.556 & 0.182     & \textbf{1.000} & 0.591 & 0.342    & \textbf{0.001} & 0.944 & 0.998      \\
FGSM\_Iter\_Target\_Person & \textbf{0.517} & 0.495 & 0.491     & \textbf{0.505} & 0.500 & 0.495    & \textbf{0.927} & 0.948 & 0.959      \\
FGSM\_Iter\_noTarget       & \textbf{0.574} & 0.497 & 0.532     & \textbf{0.544} & 0.503 & 0.531    & \textbf{0.870} & 0.950 & 0.936      \\
FGSM\_noIter\_Target\_Road & \textbf{0.848} & 0.501 & 0.224     & \textbf{0.833} & 0.495 & 0.349    & \textbf{0.510} & 0.944 & 0.992      \\
GD-UAP                     & 0.568 & 0.489 & \textbf{0.327}     & 0.534 & 0.495 & \textbf{0.386}    & \textbf{0.878} & 0.950 & 0.982      \\
\bottomrule
\end{tabular}}
\caption{Performance of the detection model for different perturbation of the test set of Cityscapes (best performing model highlighted in bold for each metric). DAAIN clearly outperforms the simple baselines on the majority of OOD inputs, but struggles to detect adversarial attacks reliably.}
\label{tab:performance}
\end{table}

\section{Conclusion \& Outlook}\label{sec:conclusion}
Our proposed method and evaluation have shown that training Normalizing Flows on a segmentation model's activations is a valuable tool to detect OOD and AA input. Using subsampling allows this method to scale to large input- and model-sizes while making it possible to detect high-impact perturbations even in the case of strong subsampling. We show that combining several approaches, such as using Normalizing Flows \cite{rudolphSameSameDifferNet2020} for density estimation, monitoring and subsampling inner activations \cite{nakka2020indirect}, exploiting the distributional differences in the activation statistics of regular and anomalous input \cite{grosseStatisticalDetectionAdversarial2017} and constructing of simple discriminative classification heads allows us to detect OOD and AA inputs in a unified framework at scale. We show that the features extracted by a Normalizing Flow give significantly outperform baselines adapted from classication despite our simple detection heads being parameter-free. This indicates that representation learning on the activation distributions is helpful. The detection model is lightweight and as such compute-efficient; it can be trained on a single GPU in a reasonable time and deployed in an accelerator-free environment, e.g., in an autonomous car.

This work constitutes a preliminary study to combine the detection of OOD and AA input for image segmentation in a single unified framework that is scalable to real-world full-scale images. Future work should focus on rigorously testing the framework against black-box adversaries as well as prove its transferability to other image segmentation settings and data modalities. This work opens up new directions to investigate detection model architectures that are more targeted towards representation learning, learning parameter-dependent classification heads as well as looking for even more compute-efficient but nevertheless robust detection models.


\section*{Author contributions, Acknowledgements and Disclosure of Funding} \label{sec:ac}

S.v.B. contributed to the design and implementation of the research, performed the experiments and analysis, and wrote the paper. J.O. contributed to the analysis design and wrote the paper. A.L. contributed to the design of the research, contributed analysis tools, and supervised the work. M.S. contributed to the design of the research and supervised the work. T. W. contributed to the design of the research and supervised the work. All authors discussed the results and contributed to the final manuscript.

This work has been funded by EU ECSEL Project SECREDAS Cyber Security for Cross Domain Reliable Dependable Automated Systems (Grant Number: 783119). The authors would like to thank the consortium for the successful cooperation.

\bibliographystyle{plainnat}
\small
\bibliography{references}

\section*{Checklist}

The checklist follows the references.  Please
read the checklist guidelines carefully for information on how to answer these
questions.  For each question, change the default \answerTODO{} to \answerYes{},
\answerNo{}, or \answerNA{}.  You are strongly encouraged to include a {\bf
justification to your answer}, either by referencing the appropriate section of
your paper or providing a brief inline description.  For example:
\begin{itemize}
  \item Did you include the license to the code and datasets? \answerYes{See Section~\ref{gen_inst}.}
  \item Did you include the license to the code and datasets? \answerNo{The code and the data are proprietary.}
  \item Did you include the license to the code and datasets? \answerNA{}
\end{itemize}
Please do not modify the questions and only use the provided macros for your
answers.  Note that the Checklist section does not count towards the page
limit.  In your paper, please delete this instructions block and only keep the
Checklist section heading above along with the questions/answers below.

\begin{enumerate}

\item For all authors...
\begin{enumerate}
  \item Do the main claims made in the abstract and introduction accurately reflect the paper's contributions and scope?
    \answerYes{}
  \item Did you describe the limitations of your work?
    \answerYes{ We address shortcomings that are relevant points throughout the paper.}
  \item Did you discuss any potential negative societal impacts of your work?
    \answerNo{No impact is to be expected.}
  \item Have you read the ethics review guidelines and ensured that your paper conforms to them?
    \answerYes{}
\end{enumerate}

\item If you are including theoretical results...
\begin{enumerate}
  \item Did you state the full set of assumptions of all theoretical results?
    \answerNA{}
	\item Did you include complete proofs of all theoretical results?
    \answerNA{}
\end{enumerate}

\item If you ran experiments...
\begin{enumerate}
  \item Did you include the code, data, and instructions needed to reproduce the main experimental results (either in the supplemental material or as a URL)?
    \answerYes{We provide a link to the code repository. However, we do not provide model weights or the code to generate the attacks. Cityscapes is licensed and cannot be redistibuted.}
  \item Did you specify all the training details (e.g., data splits, hyperparameters, how they were chosen)?
    \answerYes{But we refer to the repository: \href{https://anonymous.4open.science/r/daain}{https://anonymous.4open.science/r/daain}}
	\item Did you report error bars (e.g., with respect to the random seed after running experiments multiple times)?
    \answerNo{We did not observe strong variations in the result and left out the deviations for presentational purposes.}
	\item Did you include the total amount of compute and the type of resources used (e.g., type of GPUs, internal cluster, or cloud provider)?
    \answerYes{See Section \ref{sec:training_setup}}
\end{enumerate}

\item If you are using existing assets (e.g., code, data, models) or curating/releasing new assets...
\begin{enumerate}
  \item If your work uses existing assets, did you cite the creators?
    \answerYes{We cite ESPNet, Cityscape, FrEIA, PyTorch as assets we based our work upon}
  \item Did you mention the license of the assets?
    \answerNo{These are standard assets, publicly available. We checked that we are allowed to use them in this work.}
  \item Did you include any new assets either in the supplemental material or as a URL?
    \answerNA{}
  \item Did you discuss whether and how consent was obtained from people whose data you're using/curating?
    \answerNA{}
  \item Did you discuss whether the data you are using/curating contains personally identifiable information or offensive content?
    \answerNo{The data which we are using does not contain PII nor offensive content.}
\end{enumerate}

\item If you used crowdsourcing or conducted research with human subjects...
\begin{enumerate}
  \item Did you include the full text of instructions given to participants and screenshots, if applicable?
    \answerNA{}
  \item Did you describe any potential participant risks, with links to Institutional Review Board (IRB) approvals, if applicable?
    \answerNA{}
  \item Did you include the estimated hourly wage paid to participants and the total amount spent on participant compensation?
    \answerNA{}
\end{enumerate}

\end{enumerate}

\end{document}